\documentclass[runningheads]{llncs}

\usepackage[year=2024]{eccv}

\usepackage{eccvabbrv}
\usepackage{subcaption}

\usepackage{graphicx}
\usepackage{booktabs}
\usepackage{cite}
\usepackage{amsmath,amssymb,amsfonts}
\usepackage{algorithmic}
\usepackage{subcaption}
\usepackage{textcomp}
\usepackage{xcolor}
\usepackage{lmodern} %
\usepackage{xltabular}
\usepackage{threeparttable}
\usepackage{svg}
\usepackage{url}
\usepackage{array} %
\usepackage{makecell} %
\usepackage{siunitx} %
\usepackage[accsupp]{axessibility}  %
\usepackage[inline]{enumitem}

\usepackage{microtype}

\usepackage{multirow}

\usepackage[pagebackref,breaklinks,colorlinks,citecolor=eccvblue]{hyperref}

\usepackage{orcidlink}

\begin{document}

\title{Novel Artistic Scene-Centric Datasets for Effective Transfer Learning in Fragrant Spaces}
\titlerunning{Artistic Scene-Centric Datasets}

\author{Shumei Liu\inst{1}\orcidlink{0009-0001-7509-2275} \and
Haiting Huang\inst{1}\orcidlink{0009-0002-9159-6139} \and Mathias Zinnen\inst{1}\orcidlink{0000-0003-4366-5216} \and
Vincent Christlein\inst{1}
\orcidlink{0000-0003-0455-3799}
}

\authorrunning{Shumei Liu et al.}

\institute{Pattern Recognition Lab, Friedrich-Alexander-Universität Erlangen-Nürnberg, Erlangen, Germany\\
\email{\{shumei.liu,mathias.zinnen\}@fau.de}}

\maketitle

\begin{abstract}
  Olfaction, often overlooked in cultural heritage studies, holds profound significance in shaping human experiences and identities.
  Examining historical depictions of olfactory scenes can offer valuable insights into the role of smells in history. 
  We show that a transfer-learning approach using weakly labeled training data can remarkably improve the classification of fragrant spaces and, more generally, artistic scene depictions.
  We fine-tune Places365-pre-trained models by querying two cultural heritage data sources and using the search terms as supervision signal.
  The models are evaluated on two manually corrected test splits.
  This work lays a foundation for further exploration of fragrant spaces recognition and artistic scene classification.
  All images and labels are released as the ArtPlaces dataset at \url{https://zenodo.org/doi/10.5281/zenodo.11584328}.
  
  \keywords{Transfer learning \and Fragrant spaces \and Scene classification \and Olfaction}
\end{abstract}

\section{Introduction}
\label{sec:intro}

Olfaction is an essential sensory modality that adds richness and depth to our life experiences, influencing our perception, behavior, and well-being in profound ways. 
Smell is also deeply intertwined with cultural traditions, rituals, and practices, 
as it shapes our experiences and identities in diverse cultural contexts.
For instance, incense and fragrances have historically held pivotal positions in religious practices, medicine, and social relations across Western civilizations %
from ancient Egypt to the eighteenth century. 
Nevertheless, olfaction and its aesthetic potential have traditionally been neglected and devalued, and the human sense of smell was viewed as a redundant evolutionary remnant~\cite{shiner2020art}.
Given the prevailing emphasis on visual-centric perspectives, locating pertinent information about smell in modern digital cultural heritage remains a critical task\cite{ehrichNoseFirstOlfactoryGaze}. 

In recent years, an increasing number of scholars and institutions %
have started to challenge the ``olfactory gaze'', unveiling the significant role of smell, and pioneering novel approaches to the study of history and culture through olfaction-centric exploration\cite{bembibre2022smelly,ali2022musti,ehrichNoseFirstOlfactoryGaze,vanerpMoreNameRose2023}. 
The paradox of smell lies in its dual nature as both our most instinctive sense and one that is particularly difficult to articulate. 
Physical smell perception is inevitably absent 
when experiencing visual mediums such as artworks. 
The automatic extraction of scent-related cues thus has to rely on proxies such as smell-related objects, gestures, scenes or iconography which indirectly suggest the presence of olfactory dimensions in a painting~\cite{zinnen2021see}.
While the recognition of smell-related objects~\cite{seeingissmelling, zinnen2022transfer, odorchallenge,odordataset}, olfactory gestures~\cite{zinnen2023sniffyart}, and even emotions~\cite{patoliya2024smellemotionrecognisingemotions} have been addressed previously, the classification of fragrant spaces has not yet been explored. 

The primary challenge arises from the absence of published datasets annotated with scene information for historical artworks, particularly those specific to artistic fragrant spaces. 
Consequently, direct training for our classification tasks 
is not possible.
Numerous datasets have been curated specifically for scene classification in real-world environments, \eg, Scene15\cite{lazebnik2006beyond,oliva2001modeling}, MIT Indoor67\cite{quattoni2009recognizing}, SUN\cite{xiao2010sun}, and also multi-million-item scene-centric datasets such as Places365\cite{zhouPlaces10Million2018}. 
However, we observe a significant performance drop when we apply models trained on these large-scale photographic data to our artistic target data.
To bridge this gap, we leverage the overlap between fragrant places and the large number of scene categories annotated in Places365.
This allows us to use photographic data for pre-training while querying open heritage data to enable fine-tuning with artistic imagery. 
Using the query terms as weak labels, we create the \textit{ArtPlaces} dataset, consisting of 4623 artworks.

We evaluate our approach in two ways: 
\textit{First}, we specifically assess the model's performance in classifying fragrant spaces.
To this end, we evaluate the models using a small set of smell-related artworks \textit{Fragrant Spaces}, manually annotated with Places365 labels.
\textit{Secondly}, to achieve a more reliable evaluation for general scene classification, we extend \textit{Fragrant Spaces} with 747 images from \textit{ArtPlaces} and manually correct the weak labels to obtain a reliable test set.

To the best of our knowledge, this study represents the first attempt to identify fragrant places in artworks.
In particular, our contributions are as follows:
\begin{itemize}
    \item We construct the ArtPlaces dataset, consisting of 3804 weakly labeled artworks for training, and 975 manually labeled artworks for evaluating artistic scene recognition systems. All annotations and images are published on Zenodo.\footnote{\url{https://zenodo.org/records/11584328}}
    \item We apply four different classifier architectures and analyze their performance with respect to different pre-training, fine-tuning, and evaluation schemes.
\end{itemize}

\section{The ArtPlaces Dataset}
\label{sec:dataset}

\subsection{Pre-Training Data}

All our models are pre-trained on the standard benchmark of Places365 (\textit{Places365-Standard})~\cite{zhouPlaces10Million2018}, using the predefined splits. 
Places365 consists of 10 million photographs, annotated with 365 scene categories and serves as a standardized benchmark for evaluating scene recognition algorithms. 

\subsection{Data Collection}

\subsubsection{Artistic Scene-Centric Datasets.} We create two weakly annotated datasets by retrieving images from the APIs of the Rijksmuseum collection\footnote{\url{https://data.rijksmuseum.nl/object-metadata/api/}} and Wikidata\footnote{\url{https://query.wikidata.org/}}.
The respective query terms are used as weak (\ie semi-automatically generated) labels which serve as supervision signals during fine-tuning.
Additionally, we create a manually labeled dataset of olfaction-related artworks to test the algorithms' capability to classify fragrant spaces.

\paragraph{RASD.}
The Rijksmuseum is the national museum of the Netherlands, which provides a treasure trove of data with extensive descriptions of over 500,000 art historical objects, a vast collection of object photographs and the complete library catalogue. The Rijksmuseum offers multiple access points for its dataset. 
We use the APIs and Image requests service provided by the platform to access the structured metadata (OAI\_DC)\footnote{\url{https://data.rijksmuseum.nl/object-metadata/}} and images of the Rijksmuseum collection.   

We select and restrict metadata information such as object type, material, technique, etc. and query the images according to search terms matching the 365 Places365 category names. 
The resulting weakly labeled images comprise the Rijksmuseum Artistic Scene-Centric Dataset (RASD).

\begin{table}[t]
  \centering
  \caption{Selected Wikidata Properties}
  \label{tab0} 
  \begin{threeparttable}
  \setlength{\tabcolsep}{6pt}
  \begin{tabular}{cccc}
  \toprule
  \textbf{Property} & \textbf{Description} & \textbf{Value} \\
  \midrule
  P18 & image & image \\ 
  P31 & instance of & painting item\\
  P180 & depicts & Places365-Standard categories \\
  \bottomrule
  \end{tabular}
  \end{threeparttable}
\end{table}

\paragraph{WASD.}
The second intermediate dataset is derived from Wikidata. Wikidata serves as a freely accessible knowledge base that functions as a centralized storage facility for structured data spanning diverse fields.
It consists of 1.54 billion item statements as of early 2023\footnote{\url{https://en.wikipedia.org/wiki/Wikidata}}, and the size is continuously growing as more data is added and updated by contributors from around the world. Each item within is uniquely identified by a QID and described through its respective statements, which consist of a property and corresponding value. 

We use the MediaWiki API and the SPARQL endpoint provided by the Wikidata Query Service to query the data.
To identify depictions of specific scenes, we match the relevant QID for each Places365 category name and build SPARQL queries using properties from the WikiData ontology (\cf~\cref{tab0}).

Specifically, we associate the scene labels with the property P180, which indicates events portrayed in any form of art. The responses comprise the URLs of the desired images, facilitating their download. Finally, the Wikidata Artistic Scene-Centric Dataset (WASD) is compiled, each labeled with a scene category.

\begin{figure}[t]
  \centering
  \includegraphics[width=0.68\textwidth]{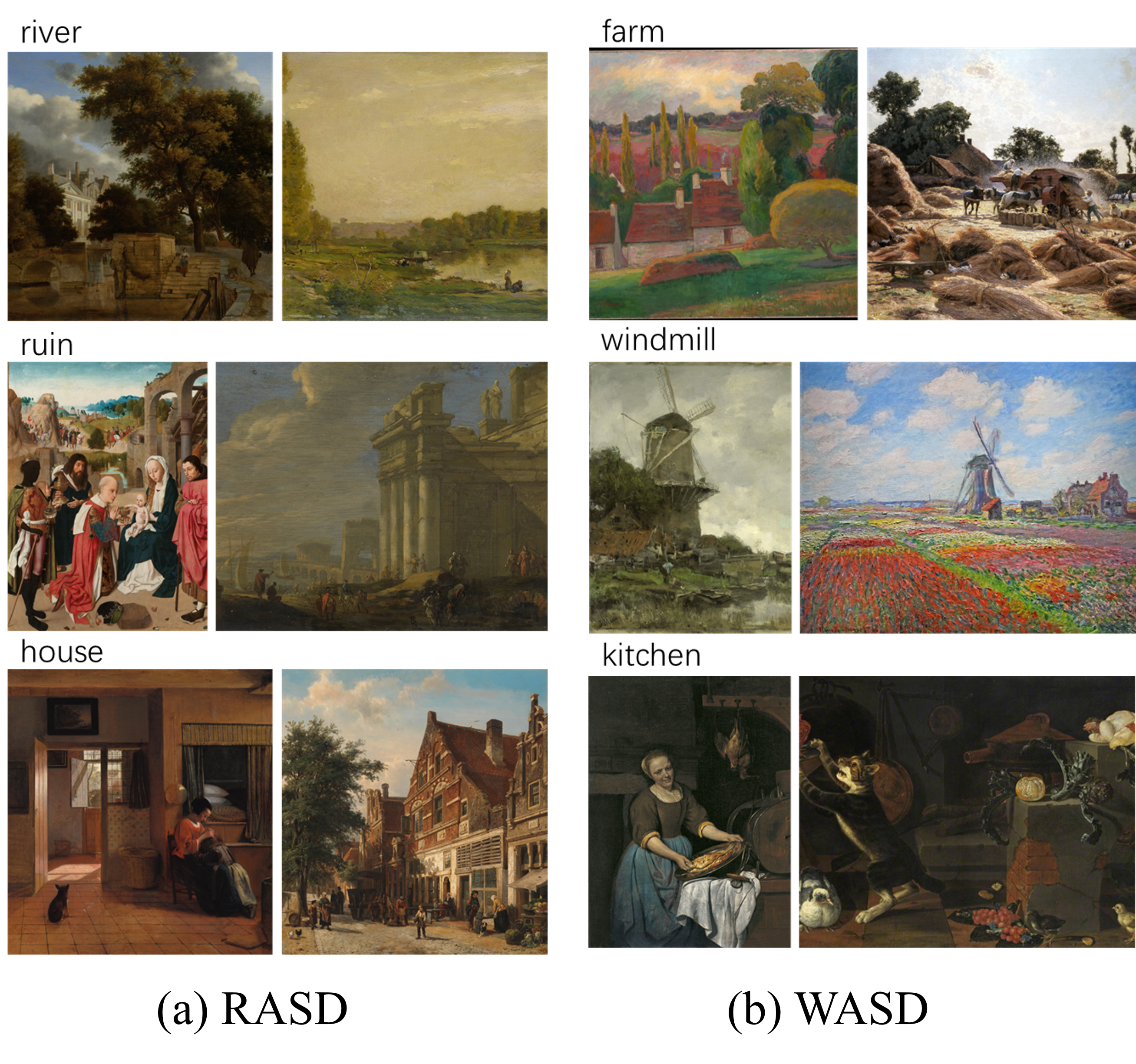}
  \caption{Image samples of the artistic scene-centric datasets.}
  \label{fig:examples}
\end{figure}

We split validation sets from both weakly labelled datasets, RASD and WASD,
to allow the analysis of models trained on only one of the subsets and a cross-dataset comparison. 
Furthermore, we transformed the multi-label problem into multiple single-label classification problems to allow comparison with single-label classification algorithms.
We provide example images for RASD and WASD in \cref{fig:examples}.

\begin{figure}[t]
  \centering
  \includegraphics[width=0.68\textwidth]{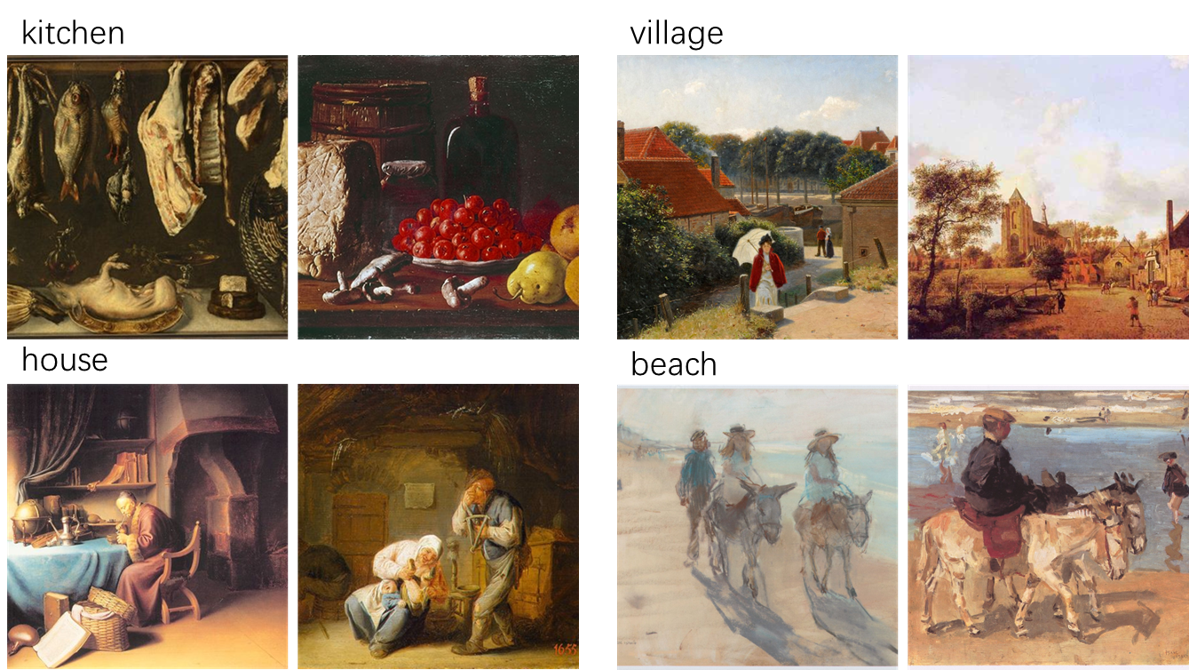}
  \caption{Image samples of the Fragrant-Spaces dataset.}
  \label{fig:fragrant_examples}
\end{figure}

\paragraph{Fragrant-Places.}
The third source subset is created by manually reviewing all artworks of the ODOR dataset~\cite{odordataset} and assigning Places365 labels where applicable. 
This \textit{Fragrant-Places} subset consists of 228 artworks and covers 35 Places365 labels.
\Cref{fig:fragrant_examples} shows exemplary image samples for some of the most common categories.


\subsubsection{Final Data Split.}
We combine the three source datasets to derive the \textit{ArtPlaces} dataset as follows: First, we combine the RASD and WASD subsets to generate a joint weakly supervised dataset. We refer to this combination as ArtPlaces-raw. From this combination, we split \SI{88}{percent} of the data for training, resulting in ArtPlaces-train. 
The labels of the remaining \SI{12}{percent} of ArtPlaces-raw were manually corrected and then merged with Fragrant-Places, resulting in ArtPlaces-test.  
Additionally, we keep a separate version of the Frag\-rant-Spaces test set.

\begin{figure}
    \centering
    \includegraphics[width=0.9\textwidth]{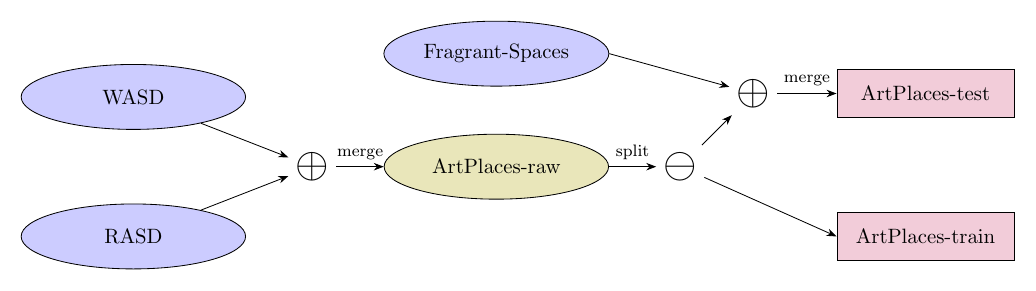}
    \caption{Illustration of the dataset creation process. ArtPlaces-raw is derived from the weakly labeled RASD and WASD subsets. From ArtPlaces-raw, ArtPlaces-train is split, and in conjunction with Fragrant-Spaces ArtPlaces-test.}
    \label{fig:subsets}
\end{figure}

This leaves us with three final data subsets:
\begin{enumerate}
    \item \emph{ArtPlaces-train}: Is the weakly labeled training split, obtained by combining parts of RASD and WASD.

\item \emph{Fragrant-Spaces}: The primary objective of the Fragrant-Spaces test set is to evaluate the models' ability to detect fragrant spaces in olfaction-related artworks.
As it is based on the ODOR dataset~\cite{odordataset}, we can assume that all of the images have some relation to olfaction. 
This focused approach allows us to measure how well the models identify and classify fragrant environments, which is crucial for their application in automated smell-reference extraction.

\item \emph{ArtPlaces-test}: In contrast, the ArtPlaces-test split is designed to provide a broader evaluation of scene classification capabilities. It does not specifically focus on olfaction-related images but aims to assess the general scene classification capabilities. 
It provides more images and covers more of the Places365 categories, thus enabling a more reliable evaluation. 
This dataset offers a robust framework for testing scene recognition performance at the expense of a specific focus on smell-related scenes. 
\end{enumerate}
\Cref{tab:subsets} gives an overview of all subsets described above and \cref{fig:subsets} illustrates their creation process.

\begin{table}[t]
    \centering
    \caption{Complete overview of the datasets and subsets used during this work. Images denote the number of images in the respective subset, coverage the number of Places365 labels covered, and labels reports on how the labels were obtained.}
    \setlength{\tabcolsep}{6pt}
    \begin{tabular}{lccc}
    \toprule
        \textbf{Dataset} & \textbf{Size} & \textbf{Coverage} & \textbf{Manual Labels} \\  
        \midrule
        RASD & \phantom{0}704 & \phantom{0}98 & \\
        WASD & 3691 & 124 & \\
        Fragrant-Spaces & \phantom{0}228 & \phantom{0}35 & \checkmark \\
        ArtPlaces-raw & 4395 & 155 & \\
        ArtPlaces-train & 3804 & 120 & \\
        ArtPlaces-test & \phantom{0}975 & 174 & \checkmark \\
        \bottomrule
    \end{tabular}
    \label{tab:subsets}
\end{table}

\subsection{Source Data Analysis}

\begin{figure}[h!t]
  \centering
  \begin{minipage}{0.95\textwidth}
      \centering
      \begin{subfigure}[b]{\textwidth}
          \includegraphics[width=\textwidth]{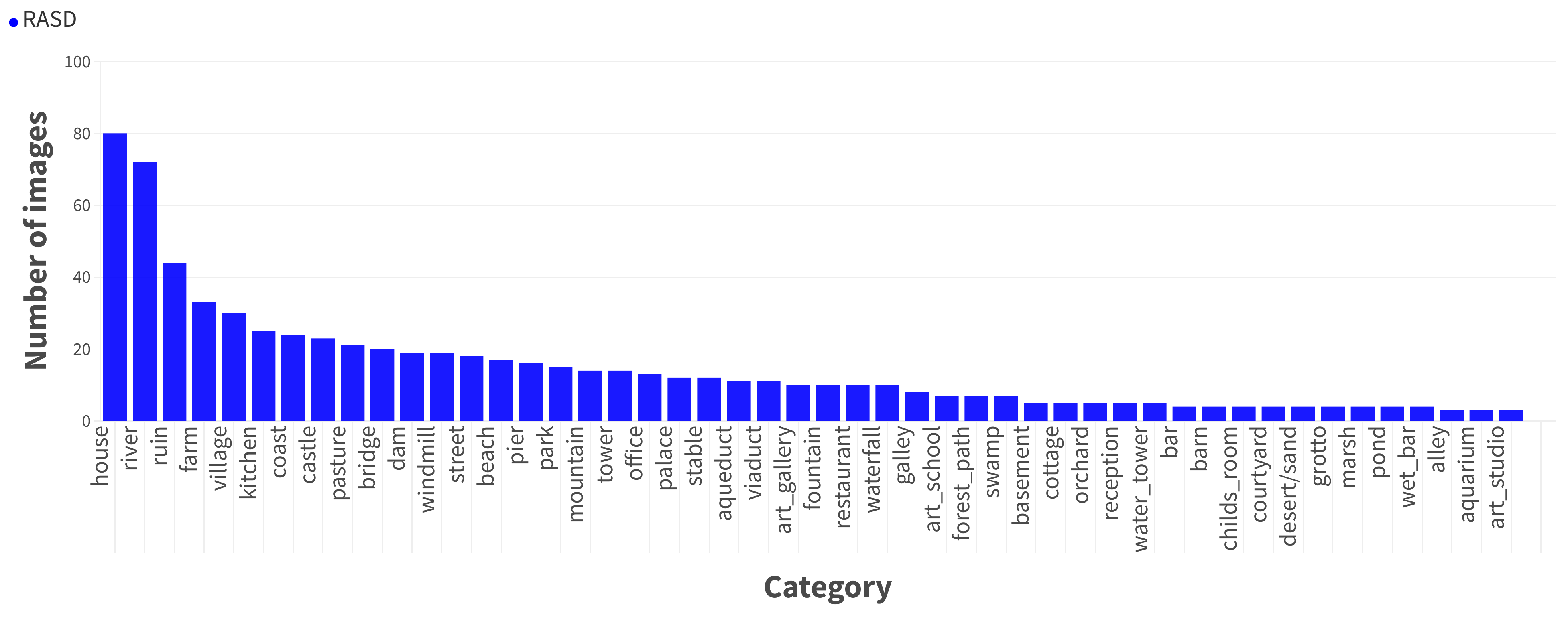}
          \caption{RASD}
          \label{fig:r}
      \end{subfigure}
  \end{minipage}
  \begin{minipage}{0.95\textwidth}
      \centering
      \begin{subfigure}[b]{\textwidth}
          \includegraphics[width=\textwidth]{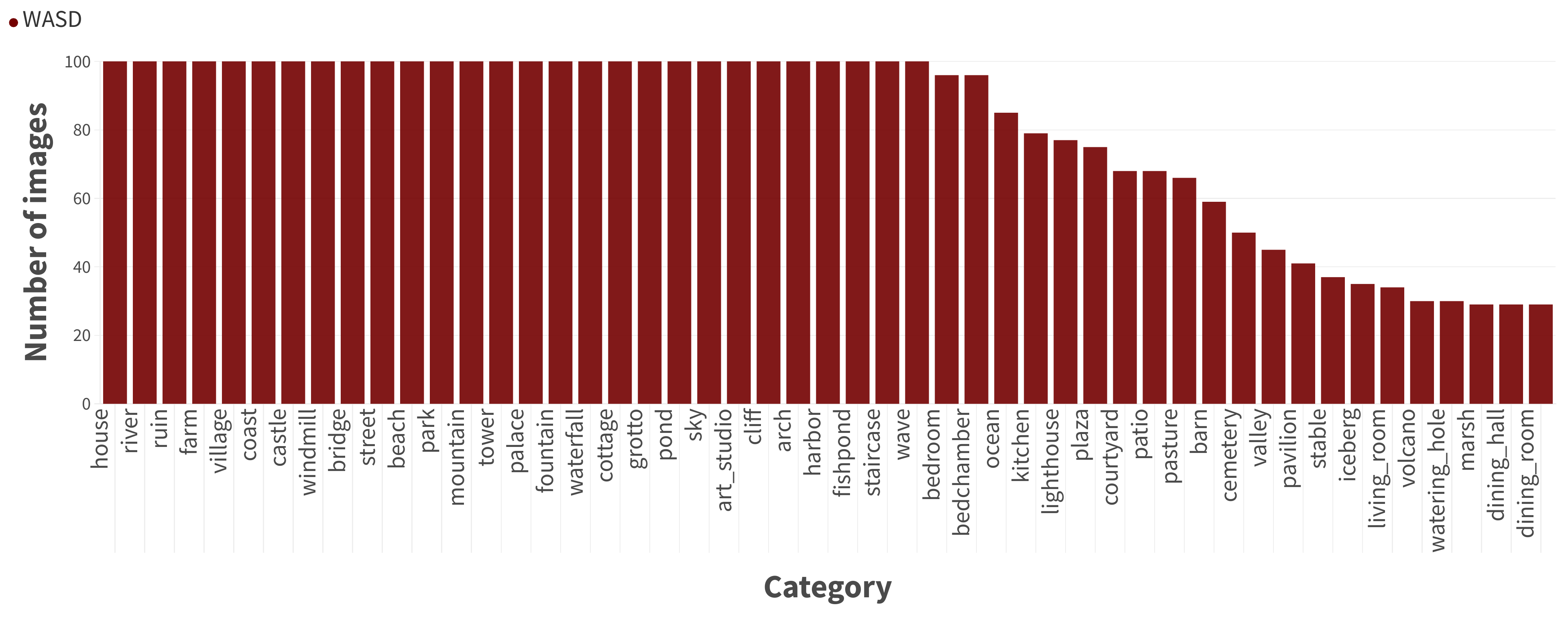}
          \caption{WASD}
          \label{fig:w}
      \end{subfigure}
  \end{minipage}
  \begin{minipage}{0.95\textwidth}
    \centering
    \begin{subfigure}[b]{\textwidth}
        \includegraphics[width=\textwidth]{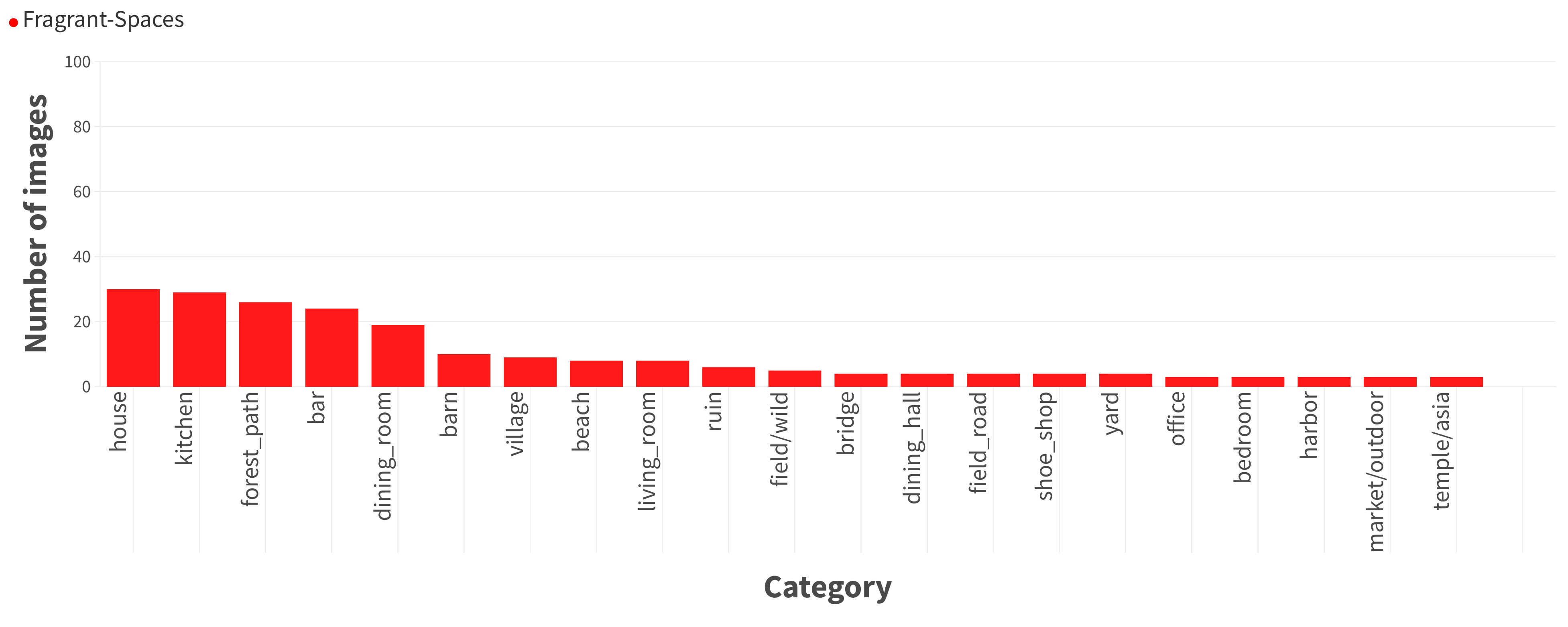}
        \caption{Fragrant-Spaces dataset}
        \label{fig:f}
    \end{subfigure}
  \end{minipage}
  \caption{Dominant categories in different datasets, contributing separately \SI{90}{\percent} of each data volume.}
  \label{fig:datasets}
\end{figure}
The source datasets are summarized in the top three columns of ~\cref{tab:subsets}. WASD has a larger collection of artworks, comprising 3691 images, compared to RASD's total of 704 pieces. Both weakly labeled datasets are considerably larger than the manually labeled Fragrant-Places subset.

\textit{Diversity.} Both weakly supervised datasets encompass a wide range of scene types, with more than one hundred categories each. \Cref{fig:datasets} indicates examples of the more frequent scenes. Although gathered from different sources, the datasets show a similar composition of scene categories. 

\textit{Multi-labeling.} We preserve the complete scene information by marking the datasets with multiple labels, effectively enabling multi-label approaches to be applied to the dataset in future work.%

\textit{Class Imbalance.} Despite capping the query results to 100 results per class, the collected data shows a significant class imbalance. 
When sorting the categories by size, the 90th percentile classes of RASD and WASD are ranked 48th and 49th separately, indicating that half of the categories encompass most of the data.

\subsection{Label Quality \& Correction}

\subsubsection{Quality Evaluation.} 
We assess the quality of our dataset samples using random sampling. Considering the size of our data, 
we decided to use stratified sampling. 
This involves the random selection of \SI{5}{\percent} samples from each class within the original datasets, ensuring that each class is represented in the sample set according to its original proportion.
Then we perform a detailed manual inspection, including visual validation, and source and background verifications.

We repeat this procedure before and after the corrections applied during the creation of ArtPlaces-test and report the results in ~\cref{tabquality}. Despite being automatically generated, WASD provides excellent quality outputs with an accuracy estimation of \SI{97.2}{\percent}, and only 6 artworks were found to have a wrong label. RASD has relatively more mistakes with original accuracy estimation being \SI{81.0}{\percent}. 
\begin{table}[tb]
  \centering
  \caption{Quality Evaluation of Artistic Scene-Centric Datasets}
  \label{tabquality} 
  \resizebox{1\linewidth}{!}{
  \begin{threeparttable}
  \setlength{\tabcolsep}{6pt}
  \begin{tabular}{ccccc}
  \toprule
  \textbf{Dataset} & \textbf{Samples} & \textbf{Mislabeled Samples}  & \textbf{Acc. Est. (Before)} & \textbf{Acc. Est. (After)} \\
  \midrule
  RASD & \phantom{0}63 & 12 & \SI{81.0}{\percent} & \SI{93.7}{\percent}\\
  WASD & 217 & \phantom{0}6 & \SI{97.2}{\percent} & \SI{97.2}{\percent} \\
  \bottomrule
  \end{tabular}
  \end{threeparttable}
  }
\end{table}

\subsubsection{Error Analysis.} 
The labeling errors can be split into five different groups:
\begin{itemize}
  \item \textit{Noisy Contextual Information.} Refers to the noisy background information that affects the context in which the label is extracted. 
  The Rijksmuseum includes documentation or references in each artwork's description, in which also appear words like ``art gallery'', ``art school'', \etc, that belong to Places365 categories and can be identified by mistake. This problem occurs mainly among RASD samples, with three cases.      
  \item \textit{Text Recognition Error.} Indicates the issue where the system misreads or misidentifies the label due to the presence of similar words or a wrong combination of multi-word labels in the text. Three RASD samples are mistakenly labeled, \eg, word ``bare'' is mistakenly extracted as the label ``bar''. 
  \item \textit{Missing references.} Wikidata offers a list of descriptive words of objects or scenes for its painting items, along with their references. We found four WASD samples labeled without clear reference clues or feasible manual identification. 
  \item \textit{Unknown Reasons.} Two sorts of mislabeling are discovered with unidentified causes. One of the Places365 categories, ``corn field'', may be mostly misassigned in RASD. All the five samples from this category are found to be misclassified. Other concerned labels are ``fishpond'' and ``pond'' in WASD. All samples related to these categories are currently tagged with both labels simultaneously. However, two cases out of them are checked probability not suitable for the specific ``fishpond''.   
  \item \textit{Inadequate Filtering.} The error arises because the filters used for the query are not strict enough, which occurs in RASD. Instead of representing an artwork, a sample corresponds to a photo of a physical object, satisfying one selected combination of object type and label filters.     

\end{itemize}

\subsubsection{Data Rectification.} 
\begin{itemize}
  \item \textit{Eliminate Noise.} RASD is relatively small, making it possible to manually filter out noisy data, \ie photos of non-artworks. 
  \item \textit{Remove Duplicates.} Scene identification is naturally a multi-label issue, therefor in our datasets exist duplicate images marked with different labels. We process a systematic cleansing of the redundant images while maintaining the multi-label information.      
  \item \textit{Exclude Problematic Label Data.} ``corn field'' is the only label experiencing a significant issue with systematic misassignment at scale. All its samples from RASD are wrongly labeled, in which exist a considerable quantity as well. In order to uphold the standards of training and analysis, we have decided to eliminate the entire category in RASD, specifically 69 label instances (while retaining the five corrected samples).  

\end{itemize}
By implementing the above strategies, we address the mislabeling concerns that have the most significant impact through manual correction. These measures are mainly applied to RASD, in which half of the mismatching issues are resolved, and the revised accuracy estimate achieves \SI{93.7}{\percent} (only four mislabeled cases found in the second round sampling). The remaining error types are more dispersed and have relatively minor impacts, and will be concerned in future work.

\section{Methodology}
\label{sec:methodology}

\subsection{Experimental Setup}
We employ the widely-used ResNet50\cite{he2015deep}, ResNet18\cite{he2015deep}, and DenseNet161\cite{huang2017densely} classification networks. 
All of these have pre-trained weights available for the Places365-Standard dataset.\footnote{\url{https://github.com/CSAILVision/places365/tree/master?tab=readme-ov-file}}

Additionally, we evaluate the recent WaveMix~\cite{jeevan2023wavemix} method. It comprises a series of self-similar and resolution-preserving WaveMix blocks, inside which a multi-level two-dimensional discrete wavelet transform (2D-DWT) is utilized for lossless down-sampling and token-mixing while maintaining image fidelity. 
The system strategically reorganizes spatial information guided by image priors, such as scale-invariance, shift-invariance, and sparseness of edges. This approach incorporates multi-resolution information, facilitating efficient learning with fewer 
parameters and layers, resulting in a rapid expansion of the receptive field compared to traditional CNN layers. 

All models are trained using cross-entropy loss with Stochastic Gradient Descent with Momentum as optimizer. 
Additionally, we apply geometric, spatial and color-based in-place data augmentation.%

\section{Results and Discussion}
We conduct three sets of experiments to evaluate 
\begin{enumerate*}[label=(\roman*)]
    \item the models' capability to recognise fragrant spaces in historical artworks,
    \item the broader scene classification in artistic depictions, and
    \item the impact of fine-tuning dataset properties on classification performance.
\end{enumerate*}

We evaluate the four baseline models described above in each experimental setup, pre-trained on Places365-standard.

\label{sec:results}

\subsection{Fragrant Spaces Recognition}

\begin{table}[t]
    \caption{Classification performance of models pre-trained on Places365 when evaluated on Places365-val and Fragrant-Places.}
    \label{tab:fragrantresults1}
    \centering
    \setlength{\tabcolsep}{6pt}
    \resizebox{\linewidth}{!}{
    \begin{tabular}{lccccccc}
    \toprule
         &  \multicolumn{3}{c}{\textbf{Places365-val}} && \multicolumn{3}{c}{\textbf{Fragrant-Places}}\\
         \cmidrule{2-4} \cmidrule{6-8}
         
         & \textbf{\textit{Top-1 acc.}} & \textbf{\textit{Top-5 acc.}} & \textbf{\textit{F}\textsubscript{1}} &&  
         \textbf{\textit{Top-1 acc.}} & \textbf{\textit{Top-5 acc.}} & \textbf{\textit{F}\textsubscript{1}} \\
         \midrule
         
         ResNet18\cite{he2015deep} & \SI{53.86}{\percent} & \SI{83.95}{\percent} & \SI{53.18}{\percent} && \phantom{0}\SI{2.63}{\percent} & \SI{12.72}{\percent} & \phantom{0}\SI{2.76}{\percent} \\
         ResNet50\cite{he2015deep}& \SI{55.09}{\percent} & \SI{85.10}{\percent} & \SI{54.37}{\percent} && \phantom{0}\SI{1.75}{\percent} & \SI{14.04}{\percent} & \phantom{0}\SI{3.18}{\percent} \\
         DenseNet161\cite{huang2017densely} & \SI{56.55}{\percent} & \SI{86.18}{\percent} & \SI{55.78}{\percent} && \phantom{0}\SI{4.39}{\percent} & \SI{17.54}{\percent} & \phantom{0}\SI{5.13}{\percent} \\
         WaveMix\cite{jeevan2023wavemix} & \SI{51.87}{\percent} & \SI{82.51}{\percent} & \SI{51.04}{\percent} && \phantom{0}\SI{2.63}{\percent} & \SI{18.42}{\percent} & \phantom{0}\SI{2.85}{\percent}\\
         \bottomrule
    \end{tabular}
    }
\end{table}

We measure the capability to recognize fragrant spaces in historical artworks using the Fragrant-Spaces subset of ArtPlaces.
When we directly apply models pre-trained on Places365 to the Fragrant-Spaces test set, we see a large performance decrease as listed in \cref{tab:fragrantresults1}.
Compared to the performance levels exhibited by the pre-trained models on the Places365 validation set, the classification accuracies in our target domain are extremely low, with top-5 accuracies falling below \SI{20}{\percent}, and top-1 accuracies even less than \SI{5}{\percent}. This stark contrast underscores a substantial disparity in performance between the scene classification for real-world images and that for artistic fragrant spaces. 
The pre-trained models, typically trained on large-scale datasets of real-world images, are not well-suited for capturing the nuances of artistic or abstract spaces. 
Artistic depictions of fragrant spaces may contain elements deviating from typical features seen in real-world scenes. These spaces could include abstract compositions, unconventional lighting, non-representational forms, and other artistic elements that may not be adequately represented in the learned features of a conventional scene classification model.

\begin{table}[]
    \caption{Fragrant-Spaces classification performance of models with and without fine-tuning on ArtPlaces-train.}
    \label{tab:fragrantresults2}
    \setlength{\tabcolsep}{6pt}
    \centering
    \resizebox{\linewidth}{!}{
    \begin{tabular}{lccccc}
    \toprule
       \textbf{Model} & \textbf{Fine-Tuning} & \textbf{Top-1 acc.} & \textbf{Top-5 acc.} & \textbf{F\textsubscript{1}(macro)} & \textbf{F\textsubscript{1}(weighted)}\\
       \midrule
       ResNet18\cite{he2015deep}  & & \phantom{0}\SI{2.63}{\percent} & \SI{12.72}{\percent} & \SI{2.76}{\percent} &\phantom{0}\SI{3.91}{\percent}\\
       ResNet18\cite{he2015deep} & \checkmark & \SI{18.86}{\percent} & \SI{43.42}{\percent} & \SI{6.40}{\percent}&\SI{16.48}{\percent}\\
       ResNet50\cite{he2015deep} & & \phantom{0}\SI{1.75}{\percent} & \SI{14.04}{\percent} & \SI{3.18}{\percent} &\phantom{0}\SI{2.02}{\percent}\\
       ResNet50\cite{he2015deep} & \checkmark & \SI{19.74}{\percent} & \SI{42.11}{\percent} & \SI{8.97}{\percent}&\SI{18.01}{\percent} \\
       DenseNet161\cite{huang2017densely} & & \phantom{0}\SI{4.39}{\percent} & \SI{17.54}{\percent} & \SI{5.13}{\percent} &\phantom{0}\SI{5.42}{\percent}\\
      DenseNet161\cite{huang2017densely} & \checkmark & \textbf{24.12}\,\textbf{\%} & \textbf{46.93}\,\textbf{\%} & \textbf{9.79}\,\textbf{\%}&\textbf{20.91}\,\textbf{\%} \\
       WaveMix\cite{jeevan2023wavemix} & & \phantom{0}\SI{2.63}{\percent} & \SI{18.42}{\percent} & \SI{2.85}{\percent} &\phantom{0}\SI{3.13}{\percent}\\
       WaveMix\cite{jeevan2023wavemix} & \checkmark & \SI{16.23}{\percent} & \SI{42.98}{\percent} & \SI{7.76}{\percent}&\SI{16.19}{\percent} \\
       \bottomrule
    \end{tabular}
    }
\end{table}

However, after fine-tuning with our weakly supervised ArtPlaces-train dataset, all models display an immense performance increase.
\Cref{tab:fragrantresults2} shows that all four models fine-tuned on the larger combined dataset exhibit an increase in all metrics. %
These promising outcomes highlight the effectiveness of this methodology in improving performance in artistic fragrant spaces. 
The fine-tuned DenseNet161\cite{huang2017densely} outperforms other models, providing the highest top-1 accuracy of \SI{24.12}{\percent} and top-5 accuracy of \SI{46.93}{\percent}.

Similarly, after fine-tuning, the macro-average F\textsubscript{1} scores increase by \SI{4.9}{\percent} in average. 
The weighted average F\textsubscript{1}(micro) scores improve by an even larger margin of \SI{12.3}{\percent} in average. 
In our context, the weighted-average evaluation leads to better results, because it takes class imbalance into the calculation, by considering the performance of each class but placing higher weight on classes with more samples. The outcomes of larger classes have a greater influence on the overall average.

Even though the finetuned models have seen a great performance improvement, there are still elements preventing further increases in accuracy. Unlike the training set, Fragrant-Space is single-labeled, which can exclude diverse information and reduce the hit rate. For example, an artwork named ``Wooded landscape with herdsmen and their flock crossing shallow water''
describes woods, a pond, and a path, but a single label only covers the last one, which however rarely appears in ArtPlaces-train dataset and therefore is hard to learn. The two largest categories that were significantly misclassified, ``forest path'' and ``bar'', are both those with insufficient coverage in the training set. Artistic scenes can be inherently ambiguous, because of their unreality and abstraction. Food is an important theme in smell-related artworks, in still life paintings frequently present a full table of fruits, vegetables, and utensils. It is hard to say whether this extracted fragmentary scene belongs to a dining room or a kitchen. Another limitation is manual labels can also make errors. Some scenes in historical artworks are not easy to visually identify, \eg, a goldsmith's workshop was mistakenly labeled as a ``bar''. 

\begin{figure}[t]
  \centering
  \begin{subfigure}[b]{0.3\linewidth}
      \centering
      \includegraphics[width=\textwidth]{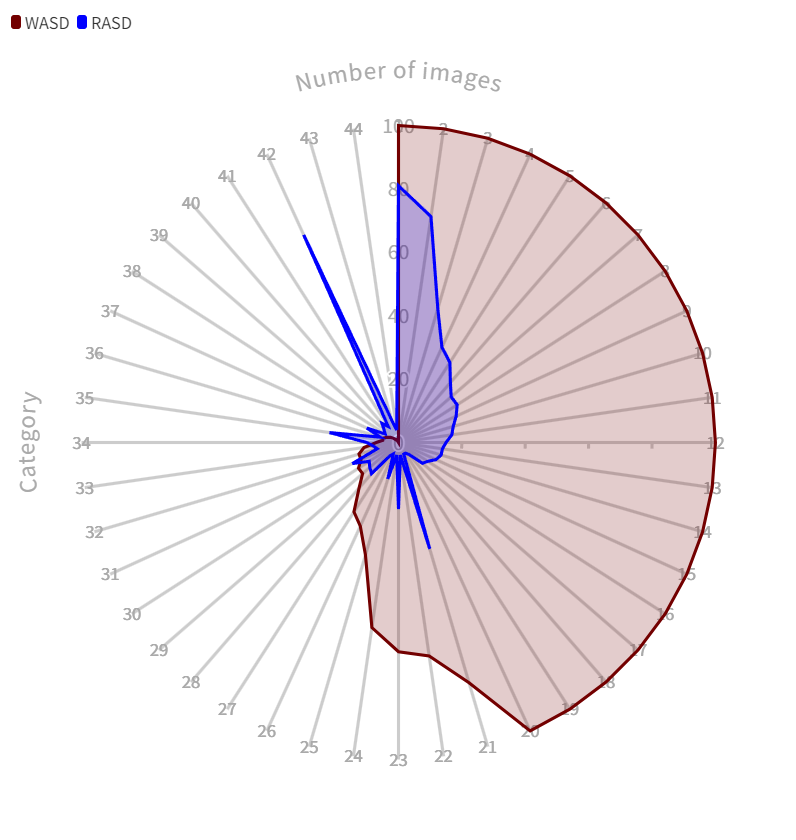}
      \caption{Based on dominant categories of RASD}
      \label{fig:rw}
  \end{subfigure}
  \hfill
  \begin{subfigure}[b]{0.3\linewidth}
      \centering
      \includegraphics[width=\textwidth]{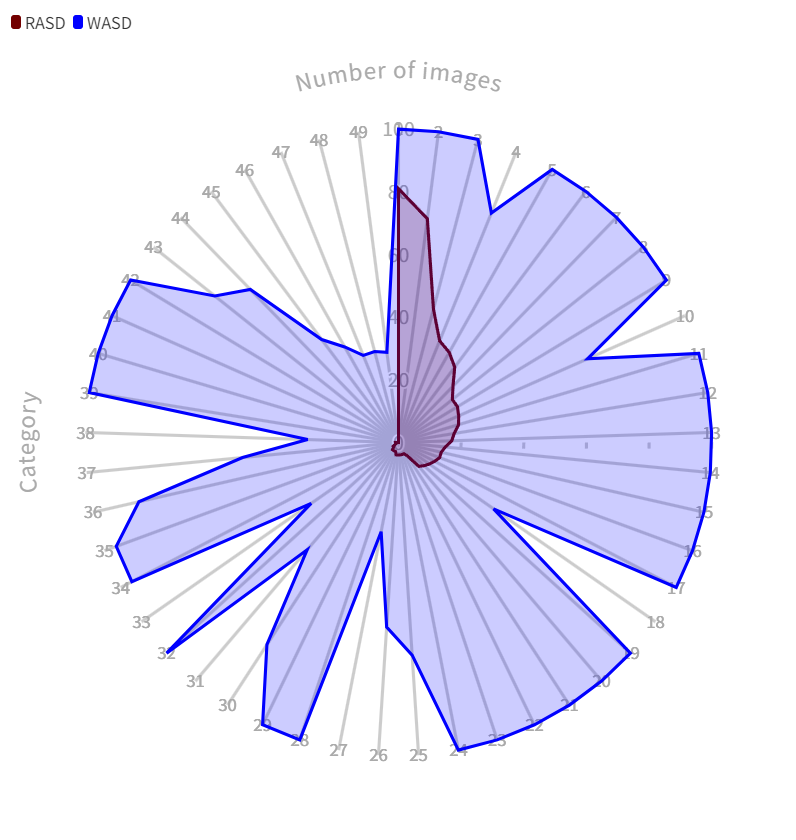}
      \caption{Based on dominant categories of WASD}
      \label{fig:wr}
  \end{subfigure}
  \hfill
  \label{fig:both}
  \begin{subfigure}[b]{.3\linewidth}
  \includegraphics[width=\textwidth]{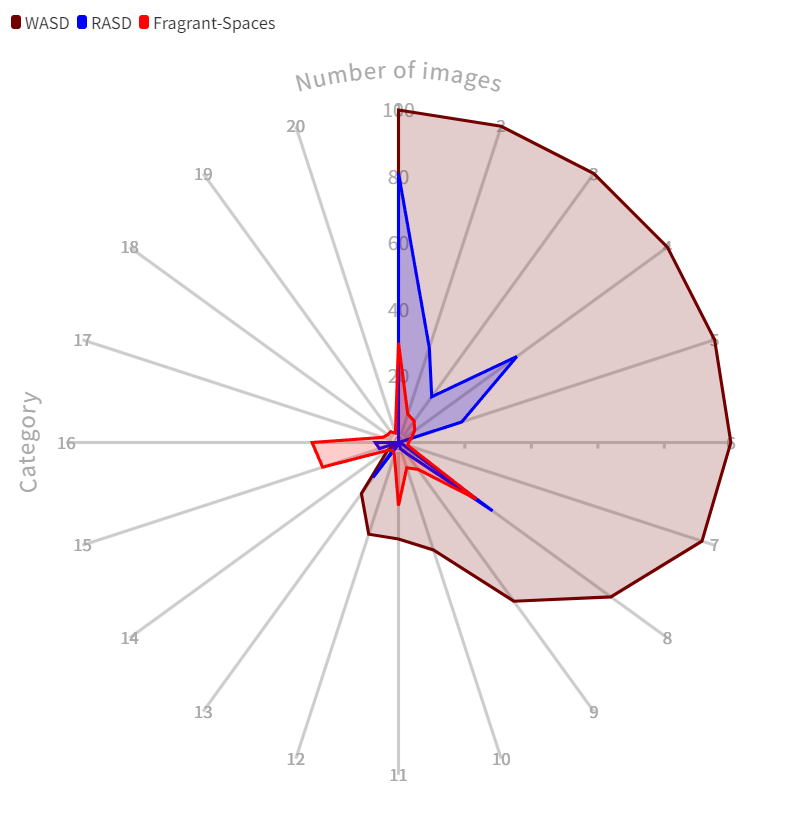}
  \caption{Based on the dominant categories of Fragrant-Spaces}
  \label{fwr}
  \end{subfigure}
  \caption{Comparison of the number of images per category between datasets RASD, WASD, and Fragrant-Spaces. In each polar coordinate, the radial axis measures the number of images per category, and the angular axis represents the sorted dominant categories.}
\end{figure}

\subsection{Artistic Scene Recognition}
We measure the general scene recognition capability using the larger ArtPlaces-test subset, which contains a broader range of images than Fragrant-Places. 
The pre-trained models are still displaying low accuracies when tasked with classifying general artistic scenes, as shown in~\cref{tab:artplacesresults2} . But both the pre-trained and fine-tuned models demonstrate increased performance on the ArtPlaces-test dataset in comparison to the smaller Fragrant-Spaces test set(\cf~\cref{tab:artplacesresults3}). 

ArtPlaces-test contains a much larger number of categories compared to ArtPlaces-train, because of the manually modified and introduced labels. Categories that do not appear in the training set cannot be identified. However, incorporating a diverse range of labels in the test set improves its representativeness, leading to more effective evaluation. The mature DenseNet structures exhibit notably enhanced accuracy rates, with a top-1 accuracy of \SI{29.73}{\percent} and a top-5 accuracy of \SI{57.40}{\percent}.

\begin{table}[t]
    \caption{Classification performance of models pre-trained on Places365 when evaluated on Places365-val and ArtPlaces-test.}
    \label{tab:artplacesresults2}
    \centering
    \setlength{\tabcolsep}{6pt}
    \resizebox{\linewidth}{!}{
    \begin{tabular}{lccccccc}
    \toprule
         &  \multicolumn{3}{c}{\textbf{Places365-val}} && \multicolumn{3}{c}{\textbf{ArtPlaces-test}}\\
         \cmidrule{2-4} \cmidrule{6-8}
         
         & \textbf{\textit{Top-1 acc.}} & \textbf{\textit{Top-5 acc.}} & \textbf{\textit{F}\textsubscript{1}} &&  
         \textbf{\textit{Top-1 acc.}} & \textbf{\textit{Top-5 acc.}} & \textbf{\textit{F}\textsubscript{1}} \\
         \midrule
         ResNet18\cite{he2015deep} & \SI{53.86}{\percent} & \SI{83.95}{\percent} & \SI{53.18}{\percent} && \phantom{0}\SI{6.95}{\percent} & \SI{25.17}{\percent} & \phantom{0}\SI{5.58}{\percent} \\
         ResNet50\cite{he2015deep} & \SI{55.09}{\percent} & \SI{85.10}{\percent} & \SI{54.37}{\percent} && \phantom{0}\SI{8.31}{\percent} & \SI{28.02}{\percent} & \phantom{0}\SI{6.41}{\percent} \\
         DenseNet161\cite{huang2017densely} & \SI{56.55}{\percent} & \SI{86.18}{\percent} & \SI{55.78}{\percent} && \phantom{0}\SI{8.77}{\percent} & \SI{29.73}{\percent} & \phantom{0}\SI{7.16}{\percent} \\
         WaveMix\cite{jeevan2023wavemix} & \SI{51.87}{\percent} & \SI{82.51}{\percent} & \SI{51.04}{\percent} && \phantom{0}\SI{9.23}{\percent} & \SI{29.73}{\percent} & \phantom{0}\SI{5.75}{\percent}\\
         \bottomrule
    \end{tabular}
    }
\end{table}

\begin{table}[]
    \caption{ArtPlaces-test classification performance of models with and without fine-tuning on ArtPlaces-train.}
    \label{tab:artplacesresults3}
    \setlength{\tabcolsep}{6pt}
    \centering
    \resizebox{\linewidth}{!}{
    \begin{tabular}{lccccc}
    \toprule
       \textbf{Model} & \textbf{Fine-Tuning} & \textbf{Top-1 acc.} & \textbf{Top-5 acc.} & \textbf{F\textsubscript{1}(macro)} & \textbf{F\textsubscript{1}(weighted)}\\
       \midrule
        ResNet18\cite{he2015deep}  & & \phantom{0}\SI{6.95}{\percent} & \SI{25.17}{\percent} & \phantom{0}\SI{5.58}{\percent} &\phantom{0}\SI{8.13}{\percent}\\
       ResNet18\cite{he2015deep} & \checkmark & \SI{23.12}{\percent} & \SI{50.23}{\percent} & \phantom{0}\SI{8.49}{\percent}&\SI{18.82}{\percent}\\
       ResNet50\cite{he2015deep} & & \phantom{0}\SI{8.31}{\percent} & \SI{28.02}{\percent} & \phantom{0}\SI{6.41}{\percent} &\phantom{0}\SI{9.05}{\percent}\\
       ResNet50\cite{he2015deep} & \checkmark & \SI{26.99}{\percent} & \SI{55.24}{\percent} & \SI{11.98}{\percent}&\SI{22.63}{\percent} \\
       DenseNet161\cite{huang2017densely} & & \phantom{0}\SI{8.77}{\percent} & \SI{29.73}{\percent} & \phantom{0}\SI{7.16}{\percent} &\SI{10.04}{\percent}\\
        DenseNet161\cite{huang2017densely} & \checkmark & \textbf{29.73}\,\textbf{\%} & \textbf{57.40}\,\textbf{\%} & \textbf{12.39}\,\textbf{\%}&\textbf{24.82}\,\textbf{\%} \\
       WaveMix\cite{jeevan2023wavemix} & & \phantom{0}\SI{9.23}{\percent} & \SI{29.73}{\percent} & \phantom{0}\SI{5.75}{\percent} &\SI{10.72}{\percent}\\
       WaveMix\cite{jeevan2023wavemix} & \checkmark & \SI{24.26}{\percent} & \SI{52.73}{\percent} & \SI{11.64}{\percent}&\SI{20.83}{\percent} \\
       \bottomrule
    \end{tabular}
    }
\end{table}

\subsection{Subset Performance Analysis}
To analyze the impact of the fine-tuning datasets, we evaluate the performance of models trained on only one of the respective subsets and perform a cross-dataset evaluation.

\begin{table}[t]
\caption{Performance analysis relative to fine-tuning with the different source subsets.}
\begin{subtable}{\textwidth}
    \caption{Classification performance on Fragrant-Spaces of models fine-tuned on RASD and WASD, respectively.}
    \label{tab:crossdatasetresults1}
    \setlength{\tabcolsep}{6pt}
    \centering
    \resizebox{\linewidth}{!}{
    \begin{tabular}{lccccccc}
    \toprule
        \multirow{2}{*}{\textbf{Model}} & \multicolumn{3}{c}{\textbf{Fine-tuned on RASD}} && \multicolumn{3}{c}{\textbf{Fine-tuned on WASD}} \\
        \cmidrule{2-4} \cmidrule{6-8}
         & \textbf{\textit{Top-1 acc.}}& \textbf{\textit{Top-5 acc.}} & \textbf{\textit{F\textsubscript{1}}} && \textbf{\textit{Top-1 acc.}}& \textbf{\textit{Top-5 acc.}} & \textbf{\textit{F\textsubscript{1}}} \\
        \midrule
        ResNet18\cite{he2015deep} & \SI{18.86}{\percent} & \SI{39.91}{\percent} &\phantom{0}\SI{4.61}{\percent} & & \SI{19.74}{\percent} & \SI{35.53}{\percent} &\SI{3.38}{\percent}\\
        ResNet50\cite{he2015deep} & \SI{18.86}{\percent} & \SI{38.60}{\percent} &\phantom{0}\SI{6.76}{\percent} & & \SI{20.18}{\percent} & \SI{39.47}{\percent} &\SI{4.70}{\percent} \\
        DenseNet161\cite{huang2017densely} & \SI{28.95}{\percent} & \SI{44.74}{\percent} &\SI{10.13}{\percent} & & \SI{19.74}{\percent} & \SI{42.11}{\percent} & \SI{5.39}{\percent}\\
        WaveMix\cite{jeevan2023wavemix} & \SI{17.98}{\percent} & \SI{40.79}{\percent} &\phantom{0}\SI{5.04}{\percent} & & \SI{14.91}{\percent} & \SI{36.40}{\percent} &\SI{3.22}{\percent}\\
        \bottomrule
    \end{tabular}
    }
\end{subtable}
\begin{subtable}{\textwidth}
    \centering
    \caption{Classification accuracy on dataset RASD}
    \label{tab3} 
    \setlength{\tabcolsep}{6pt}
    \begin{threeparttable}
    \begin{tabular}{ccccc}
    \toprule
    \multirow{2}{*}{\textbf{Model}} & \multicolumn{2}{c}{\textbf{Pretrained}} & \multicolumn{2}{c}{\textbf{Fine-tuned on WASD}}\\
    \cmidrule(lr){2-3} \cmidrule(lr){4-5} 
    \textbf{ } & \textbf{\textit{Top-1 acc.}}& \textbf{\textit{Top-5 acc.}}& \textbf{\textit{Top-1 acc.}}& \textbf{\textit{Top-5 acc.}} \\
    \midrule
    ResNet18\cite{he2015deep} & \phantom{0}\SI{5.24}{\percent} & \SI{19.64}{\percent} & \SI{15.73}{\percent} & \SI{37.27}{\percent} \\
    ResNet50\cite{he2015deep} & \phantom{0}\SI{6.29}{\percent} & \SI{20.02}{\percent} & \SI{17.73}{\percent} & \SI{40.32}{\percent} \\
    DenseNet161\cite{huang2017densely} & \phantom{0}\SI{6.48}{\percent} & \SI{20.97}{\percent} & \SI{18.02}{\percent} & \SI{38.80}{\percent} \\
    WaveMix\cite{jeevan2023wavemix} & \phantom{0}\SI{7.34}{\percent} & \SI{18.88}{\percent} & \SI{20.88}{\percent} & \SI{43.57}{\percent} \\
    \bottomrule
    \end{tabular}
    \end{threeparttable}
\end{subtable}
\begin{subtable}{\textwidth}
    \centering
    \caption{Classification accuracy on dataset WASD}
    \label{tab4} 
    \begin{threeparttable}
    \setlength{\tabcolsep}{6pt}
    \begin{tabular}{ccccc}
    \toprule
    \multirow{2}{*}{\textbf{Model}} & \multicolumn{2}{c}{\textbf{Pretrained}} & \multicolumn{2}{c}{\textbf{Fine-tuned on RASD}}\\
    \cmidrule(lr){2-3} \cmidrule(lr){4-5} 
    \textbf{ } & \textbf{\textit{Top-1 acc.}}& \textbf{\textit{Top-5 acc.}}& \textbf{\textit{Top-1 acc.}}& \textbf{\textit{Top-5 acc.}} \\
    \midrule
    ResNet18\cite{he2015deep} & \SI{10.07}{\percent} & \SI{28.92}{\percent} & \phantom{0}\SI{9.61}{\percent} & \SI{28.26}{\percent} \\
    ResNet50\cite{he2015deep} & \SI{10.92}{\percent} & \SI{30.87}{\percent} & \SI{12.68}{\percent} & \SI{33.71}{\percent} \\
    DenseNet161\cite{huang2017densely} & \SI{11.37}{\percent} & \SI{31.56}{\percent} & \SI{14.60}{\percent} & \SI{36.02}{\percent} \\
    WaveMix\cite{jeevan2023wavemix} & \phantom{0}\SI{9.86}{\percent} & \SI{28.08}{\percent} & \phantom{0}\SI{9.18}{\percent} & \SI{26.27}{\percent} \\
    \bottomrule
    \end{tabular}
    \end{threeparttable}
\end{subtable}
\end{table}

\Cref{tab:crossdatasetresults1} compares the classification performance on Fragrant-Spaces relative to the fine-tuning dataset used. 
Despite the size disparity between the RASD and WASD datasets, the performances of models fine-tuned on them are relatively similar. Their top-1 accuracies are also very close to those of models fine-tuned on the combined ArtPlaces-train set. But the top-5 accuracies are not as high compared to the latter.

However, when testing on each other, the outcomes demonstrate noticeable differences.
Models fine-tuned on RASD and evaluated on WASD exhibited significant enhancements, with an average increase of \SI{20.11}{\percent} and \SI{11.75}{\percent} in top-5 and top-1 accuracy, respectively (\cf\cref{tab3}). By contrast, when fine-tuned on WASD and tested on RASD, their performances remain relatively consistent before and after fine-tuning (\cf\cref{tab4}). 

To investigate the underlying cause of this phenomenon, we 
conduct a detailed analysis of the dataset compositions. 

As shown in~\cref{fig:rw}, WASD encompasses most of the dominant categories (\ie, including \SI{90}{\percent} data) in RASD, and contains a fair amount of data for training. 
Conversely, RASD includes about three-quarters of the dominant categories in WASD (\cf~\cref{fig:wr}), moreover 
only \SI{18}{\percent} of these categories possess more than 20 images, which is too limited compared to the size of the test set. The insufficient availability of representative training data in this scenario hinders the potential performance improvements.

We further compare all three datasets (\cf~\cref{fwr}) and find that RASD and WASD both cover the majority of the top 20 categories in the small Fragrant-Spaces set (RASD \SI{65}{\percent}, WASD \SI{75}{\percent}), and they are also complementary, providing a combined coverage of \SI{80}{\percent} along with a sufficient quantity of instances.
Accordingly, models fine-tuned on the merged dataset exhibit a large performance boost, emphasizing the crucial role of size and diversity of fine-tuning datasets.

\section{Conclusion}
\label{sec:conclusion}
This work demonstrates that both fragrant places and general artistic scene depictions can effectively be recognized using a weakly labeled transfer learning approach. 
We based our method on two open cultural heritage data sources: The Rijksmuseum and Wikidata.  
We constructed a fine-tuning training set of 3,804 artworks using the query terms as weak labels. 
To evaluate the model's ability to recognize fragrant spaces we manually labeled images from the ODOR dataset~\cite{odordataset} with Places365 labels. 
Additionally, we combined this Fragrant Places test set with a manually corrected split from the queried artworks to evaluate broader artistic scene classification performance.

A cross-dataset analysis suggests that successful transfer training requires the availability of fine-tuning datasets with sufficient representative categories and instances. 

However, compared to natural image datasets such as Places365~\cite{zhouPlaces10Million2018} or SUN~\cite{xiao2010sun} our dataset is small. 
Expanding the dataset by adding additional data sources, effectively broadening the scope of covered categories holds significant potential for future work. 
Our results show that even around 5,000 artworks lead to a notable increase in performance.
We expect this improvement to become more pronounced when scaling up the approach by processing more data sources. 
A dataset quality assessment revealed high variability in the quality of the weak labels: While the Wikidata subset achieved a high-quality level, there is room for improvement in the Rijksmuseum labels. 
Conducting similar analyses for additional data sources will be beneficial, allowing corrective measures where necessary.
This work opens up the space for additional research efforts to explore methods to advance the classification of olfactory spaces in artworks, and possibly broaden the scope to achieve a more general classification of artistic scene depictions.

\newpage
\bibliographystyle{splncs04}
\bibliography{main}
\end{document}